\documentclass[letterpaper, 10 pt, conference]{ieeeconf}

\IEEEoverridecommandlockouts % This command is only needed to use \thanks command
\usepackage{multicol}
\usepackage[bookmarks=true]{hyperref}
\usepackage[noadjust]{cite}

\usepackage{booktabs}
\usepackage{graphicx}
\usepackage{float}
\usepackage{amsmath, amssymb}
\usepackage{amsfonts}
\usepackage{bm}
\usepackage{algpseudocode}
\usepackage{algorithm}
\usepackage{multirow}
\usepackage[skip=1pt,font=small,labelfont=bf]{caption}
\usepackage{rotating}
\usepackage{subcaption}
\usepackage{stfloats}

\newcommand{\shrinka}{\def\baselinestretch{0.993}\large\normalsize}
\usepackage{amsmath}

\IEEEoverridecommandlockouts %needed to make affiliations appear

\hypersetup{
    colorlinks=true,    
    linkcolor=black,
    urlcolor=blue,
}
\usepackage{xargs}
\usepackage[pdftex,dvipsnames]{xcolor}
\usepackage[colorinlistoftodos,prependcaption,textsize=footnotesize]{todonotes}
\def \rebeccaColor {RoyalPurple}
\def \tuckerColor {PineGreen}
\def \martinColor {Blue}
\newcommandx{\rebecca}[1]{\todo[linecolor=\rebeccaColor,backgroundcolor=\rebeccaColor!25,bordercolor=\rebeccaColor,inline]{\textbf{RM:} #1}}
\newcommandx{\tucker}[1]{\todo[linecolor=\tuckerColor,backgroundcolor=\tuckerColor!25,bordercolor=\tuckerColor,inline]{\textbf{TH:} #1}}
\newcommandx{\martin}[1]{\todo[linecolor=\martinColor,backgroundcolor=\martinColor!25,bordercolor=\martinColor,inline]{\textbf{MM:} #1}}

\begin{document}

% paper title
\title{Comparing Piezoresistive Substrates for Tactile Sensing
  in Dexterous Hands}

\author{
Rebecca Miles$^1$, Martin Matak$^2$, Sarah Hood$^3$, Mohanraj Devendran Shanthi$^2$,
Darrin Young$^1$,\\ and Tucker Hermans$^{2,4}$
\thanks{$^{1}$University of Utah, Department of Electrical \& Computer Engineering; {\tt rebecca.miles@utah.edu}.}%
\thanks{$^{2}$University of Utah School of Computing and Utah Robotics
  Center, Salt Lake City, UT, USA; {\tt tucker.hermans@utah.edu}.}%
\thanks{$^{3}$University of Utah, Department of Mechanical Engineering.}
\thanks{$^{4}$NVIDIA Corporation, Seattle, USA.}
}
%\maketitle

\setcounter{figure}{1}
\makeatletter
\let\@oldmaketitle\@maketitle% Store \@maketitle
\renewcommand{\@maketitle}{\@oldmaketitle% Update \@maketitle to insert...
\begin{center}
  \centering     
  \includegraphics[width=\textwidth] {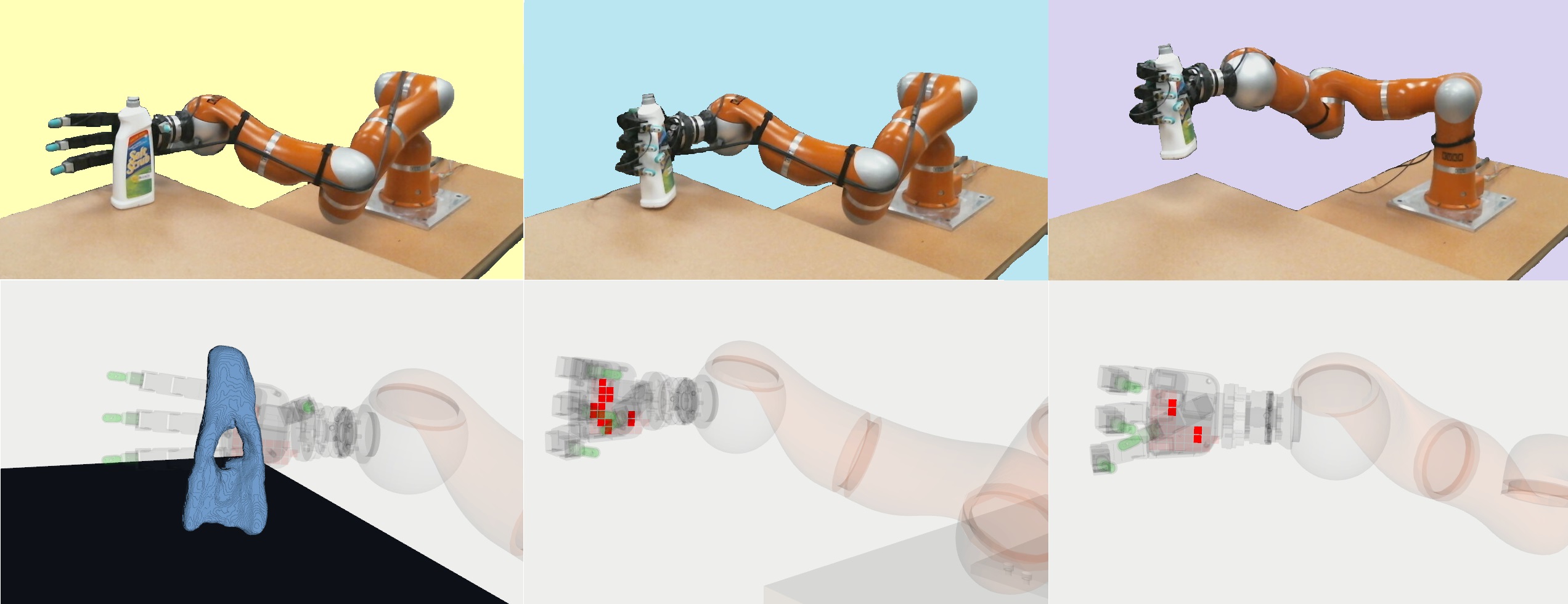}
  \label{fig:fullGrasp}
\end{center}
  %\refstepcounter{figure}
  \footnotesize{\textbf{Fig.~\thefigure:}~ Our tactile sensor
    providing feedback and analysis in three phases: (left) reaching to a grasp preshape, (center) grasping the object,
    and (right) lifting it to a desired height. The preshape is
    generated from a partial view RGB-D observation. Once the object is
    grasped, active taxels on the palm indicate contact locations
    between the hand and the object. Once lifted, a new set of taxels is
    activated, indicating a change in pressure distribution.}\vspace{-12pt}
  \medskip}% ... an image
\makeatother
\shrinka
%\maketitle
%\shrinka
\maketitle
\shrinka
\thispagestyle{empty}
\pagestyle{empty}

\newcommand{\rtax}{R_\text{tax}}
\newcommand{\rtaxk}{R_{\text{tax}_k}}
\newcommand{\rdiv}{R_\text{div}}

\begin{abstract}
While tactile skins have been shown to be useful for detecting
collisions between a robotic arm and its environment, they have not
been extensively used for improving robotic grasping and in-hand
manipulation. We propose a novel sensor design for use in covering
existing multi-fingered robot hands. We analyze the performance of
four different piezoresistive materials using both fabric and
anti-static foam substrates in benchtop experiments. We find that although the
piezoresistive foam was designed as packing material and not for use
as a sensing substrate, it performs comparably with fabrics
specifically designed for this purpose.
While these results demonstrate the potential of piezoresistive
foams for tactile sensing applications, they do not fully characterize
the efficacy of these sensors for use in robot manipulation. As such,
we use a low density foam substrate to develop a scalable tactile skin that can
be attached to the palm of a robotic hand. We demonstrate several
robotic manipulation tasks using this sensor to show its ability to
reliably detect and localize contact, as well as analyze contact
patterns during grasping and transport tasks. Our project website provides details on all materials, software, and data used in the sensor development and analysis: \url{https://sites.google.com/gcloud.utah.edu/piezoresistive-tactile-sensing/}. 

%%% Local Variables:
%%% mode: latex
%%% TeX-master: "main"
%%% End:

\end{abstract}

\IEEEpeerreviewmaketitle

\section{Introduction}
\label{sec:intro}
Existing robotics research clearly establishes the benefits of tactile
sensing for manipulation~\cite{brock1988enhancing,allen1989acquisition,veiga-toh2018-slip-prediction,Calandra2017}. Researchers have consistently shown the
benefit of tactile sensing over vision-based perception alone,
particularly in the context of manipulation under uncertainty
or involving unknown or unmodeled objects~\cite{bierbaum2008potential,dragiev2011gaussian,dang-auro2013,vanhoof-ichr2015-in-hand-rl,Calandra2017,veiga-toh2018-slip-prediction}. Tactile sensing has
been shown to improve grasp execution and evaluation~\cite{dang-auro2013,murali2018,li2014,Calandra2017,wu-corl2019-mat};
predict and arrest slip~\cite{brock1988enhancing,tremblay-icra1993,ROMANO,Heyneman2013,veiga-toh2018-slip-prediction}; perform 3D object
reconstruction~\cite{allen1989acquisition,jia2010surface,yi-iros2016-active-touch} and localization~\cite{lepora-rss2009,pezzementi-icra20110haptic-slam,li2014localization}; identify
objects~\cite{lepora-rss2009,tapo-iros2012} and their
materials~\cite{hoelscher-ichr2015-tactile-recoginition} among other properties~\cite{lepora,Chu2013,ponce-wong-trhap2014}; and enable dexterous in-hand manipulation~\cite{brock1988enhancing,vanhoof-ichr2015-in-hand-rl}.
Why then, given over 30 years of active research in robotics~\cite{brock1988enhancing,allen1989acquisition,howe1989sensing},
are large-scale tactile skins still not commonplace?

% TODO: Highlight the issues with sensor design, deployment, and
% integration
In the context of manipulation, tactile sensing has primarily focused
on the use of high-fidelity sensing at distal
phalanges~\cite{biotac,lepora,padmanabha-icra2020-omnitact,gelsight,donlon-ral2018}. Fewer
works examine the use of tactile sensing throughout the
finger~\cite{piacenza-trmech2020,takktile,jentoft2014limits} or in the
palm~\cite{jentoft2014limits} typically at the loss of sensor acuity or
modalities. This reliance on sensing only at the fingertips neglects
the potential feedback from the palm and fingers, which could improve spatial coverage in many of the above use cases.

Tactile sensing skins have more commonly been employed for contact
detection along the length of robotic arms~\cite{icubskin,Bhattarcharjee2013}. Robots
can successfully use such sensors to improve arm navigation through cluttered
environments~\cite{Killpack2013}. These existing tactile skins have
been shown to be reliable and responsive to contact, but typically rely
on highly specialized components, such as conductive fabrics or
custom circuit boards, which can be difficult to
source or construct.

Furthermore, commercially available sensors can cost thousands of
dollars. These issues compound making tactile sensors prohibitively
expensive either in cost or time for most researchers to
install. The research community's lack of use further hinders
deployment in commercial applications given the lack of common methods
for using tactile sensing.
Thus, there remains a need for easily producible and affordable
tactile sensors which can cover larger portions of a robot.

To address these issues we propose a novel tactile sensor design for
use as a tactile skin on a dexterous multi-fingered hand. Our sensor design can be constructed from multiple types of
piezoresistive material enabling us to test and compare alternative substrates for tactile skins.
Additionally, our tactile elements are of smaller scale than
previously shown in order to apply them directly on the robot palm.

\begin{figure}[h]
    \vspace{-8pt}
    \centering
     \includegraphics[width=6cm ,clip,trim=0cm 0cm 0cm 0mm]{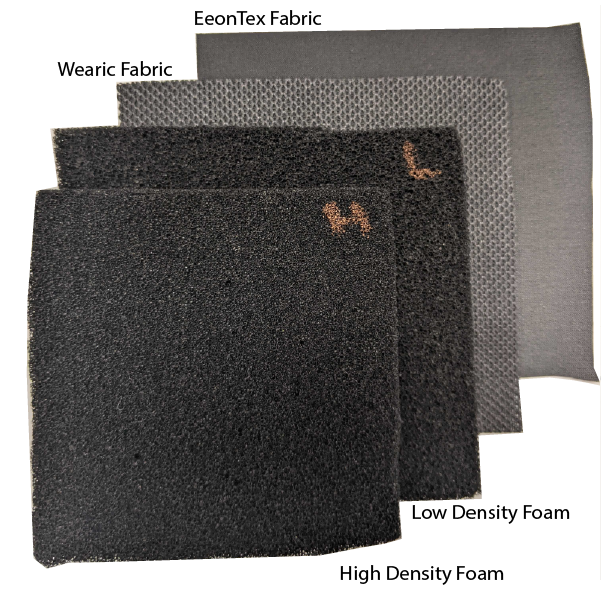} % trim order l b r t
     \caption{Piezoresistive fabric and foam substrates.}
     \label{fig:substrates}
     \vspace*{-5mm}
   \end{figure}

In this paper, we analyze the performance of two known piezoresistive
fabrics compared to two types of anti-static foam which, by nature of
their construction, exhibit piezoresistive properties. These substrates can be seen in Figure \ref{fig:substrates}.
Our motivation for using foam comes primarily as we found the fabrics
used in previous work~\cite{Day2018,Bhattarcharjee2013,Killpack2013}
to be extremely difficult to source, while foam could be easily
ordered from numerous vendors. We highlight other potential benefits of foam in
our discussion in Section~\ref{sec:conclusion}.

We report the results of highly controlled laboratory experiments
enabling us to thoroughly compare the performance of the different substrates.
In order to evaluate the effectiveness of our sensor for use in robot
manipulation, we install the sensor using our novel foam substrates on a four-fingered
robot hand attached to a 7 DOF arm. Using this palm-based
sensor we examine the efficacy of the tactile sensor to reliably
detect contact, localize the contact on the sensor array, and leverage
information from this sensor to analyze grasp and manipulation
performance. Figure~1 shows our tactile-equipped hand grasping an unknown object.

We summarize our contributions as follows: (1) we show the efficacy of using a novel sensing
substrate (foam) instead of the more difficult to source fabric for
piezoresistive tactile sensing. (2) We design an affordable, scalable tactile sensing
skin for use directly on robot hands for manipulation; (3) we
demonstrate the ability of using smaller scale tactile elements
compared with similar, previously developed sensors. (4) We 
extensively evaluate the sensor on goal-driven robot
manipulation and object recognition tasks to highlight its utility for tactile perception and feedback control. We provide explicit links to all materials, software, and data used in the sensor development and analysis as well as video at our project website: \url{https://sites.google.com/gcloud.utah.edu/piezoresistive-tactile-sensing/}. 

In the next section we provide a more detailed discussion of the
related literature. In Section~\ref{sec:sensor-design}
we discuss the sensor design and evaluate multiple piezoresistive
substrates. Section~\ref{sec:robot-experiments}
reports the details of manipulation experiments using our
sensor. We conclude in Section~\ref{sec:conclusion}.

%%% Local Variables:
%%% mode: latex
%%% TeX-master: "main"
%%% End:

\section{Related Work}
\label{sec:related-work}
In this section we discuss the use and design of tactile sensing for
robot manipulation. We primarily focus on the design aspects of
different sensing modalities used for tactile skins or in robot hands for
grasping. We then discuss the details of piezoresistive materials for
tactile and capacitive sensing that relate most closely to the sensor
proposed in this paper.

\subsection{Tactile Hands for Manipulation}
While tactile skins have become more popular in recent years, they are
primarily used as low resolution contact sensors along the length of
the robotic arm for navigating through clutter or collision avoidance
\cite{Killpack2013,Bhattarcharjee2013,Gruebele2020,Hughes2018} or on
limited regions of a robotic hand \cite{gelsight,biotac}. There has
not been a significant amount of research into using tactile skins on
robotic hands to improve grasp performance.

Tactile-enabled grasping improves success over vision-only grasping \cite{Calandra2017,murali2018,li2014}. This fact motivates robotics researchers to provide tactile sensing to robotic systems \cite{biotac,gelsight,kuppuswamy-iros2020-soft-bubble-grippers}. Fingertip sensors such as the GelSight \cite{gelsight} and BioTac \cite{biotac} measure the material distortion within the sensor when contact is made. While fingertip sensors are highly sensitive and can collect high resolution tactile information about surfaces, they neglect the potential of feedback from the remainder of the hand, namely the palm and fingers.

The most prominent existing tactile skin used in
robotic grasping is the Takktile~\cite{takktile}
array of sensors. These have been installed along the fingers and in
the palm of hands~\cite{jentoft2014limits}, including the commercially available Reflex hand. These barometric sensors are open source and can be assembled
in a matter of days for very little cost. Although they can be
manufactured in different sizes, Takktile sensors are only practical
for fine resolution sensor arrays and not for coarse large area
sensing over the entire robotic hand-arm system.

%I think I hit all of these points.
% \tucker{I think this paragraph and up until the next subsection should
%   discuss different skin sensing designs (specifically larger scale
%   designs) a bit more in terms of the technologies being used, and
%   highlight the merits of the resistive sensing design over other
%   possibilities. Motivated by this, the next subsection can focus on
%   the specific merits of foam over fabric without having to discuss
%   other sensors. For example What about the icub and its pcb skin?
%   needs to be discussed in this context}
  
Other existing tactile skins utilize capacitive \cite{Atalay2018,Gruebele2020,icubskin} or resistive \cite{Bhattarcharjee2013,Day2018,Zhao2019} properties of
materials to detect collisions. Capacitive sensors require lower power consumption compared to resistive sensors. However, as our 
sensor will always be wired to power, the difference in power consumption was 
not a significant factor to be accounted for in our design. Another cause for
the popularity of capacitive sensors derives from their high
sensitivity; Gray et al.~\cite{gray1996}
report that an 8\(\times\)8 capacitive sensing array with 1 mm$^2$ taxels
was at least ten times more sensitive than a human.

The capacitive sensors from~\cite{icubskin} enable large area contact detection along the
surface of an iCub humanoid robot. The skin is divided into a series of interconnected triangular modules, each
containing up to 192 taxels. These patches can be combined to achieve large area coverage over the entire robot. 
While the iCub tactile skin (and PCB sensors generally) can be customized to some extent for different applications,
the versatility is limited by the fixed, large size of the triangular
modules defining a patch of sensors. This restricts the ability to vary tactile
spatial resolution across the robot as needed by
the desired application.
% The taxels are also a fixed size which removes the option of
% having a higher density of taxels in one area of the robot while single large sized
% taxels are used in other areas. Having the ability to utilize varying sizes of taxel
% in different parts of the robot is helpful for collecting a large amount of information
% from the environment while minimizing the amount of data that will need to be processed. 
In contrast the piezoresistive sensors we leverage in this work scale
more easily to different sized taxels and different array shapes.
% which is better for using a single sensor design on different areas of
% the robot. For example, without changing the design, our sensor can be 
% applied in large sized taxels for collision detection along the robotic 
% arm as well as in small sized taxels for object detection on the robotic hand.
% Piezoresistive sensors are also able to detect a much wider range of pressures
% than capacitive sensors, making them a better candidate for measuring specific
% forces experienced accross the taxels. Despite there being much research into the
% use of different substrates in tactile skins, piezoresistive foam has yet to make
% its debut in the field of tactile sensing.

\subsection{Piezoresistive Materials for Tactile Skin}
\begin{figure}[ht]
    \vspace{5pt}
    \centering
    \includegraphics[trim=140 155 400 95,clip,width=0.46\textwidth]{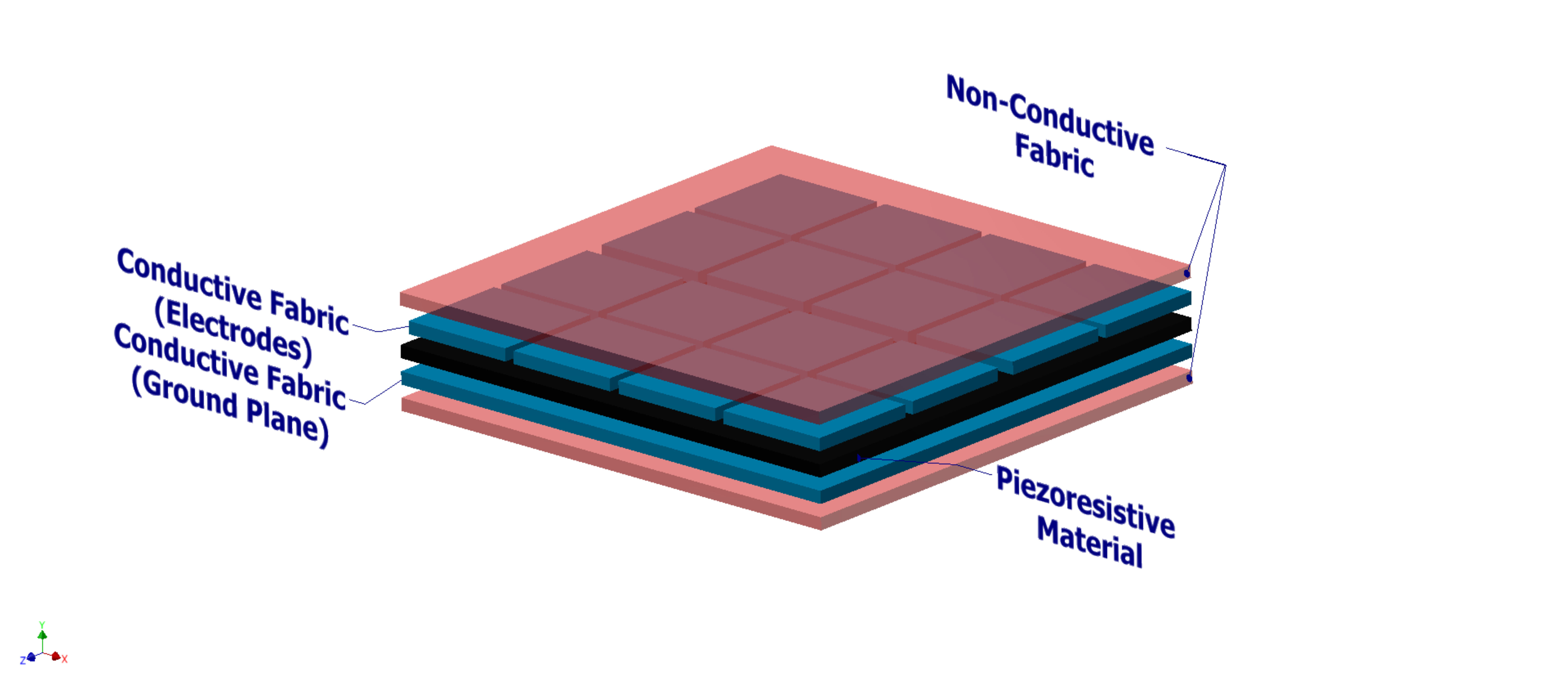}
    \caption{Five layers of non-conductive, conductive and resistive fabric that make up our tactile sensor array.}
    \label{fig:sensorDiagram}
    \vspace*{-6mm}
\end{figure}

There have been several tactile skins developed which utilize
stretchable and piezoresistive fabrics by
EeonTex~\cite{Bhattarcharjee2013,Day2018}. These sensors use a
similar design to ours where conductive fabric acts as electrodes
against the resistive substrate as illustrated in
Figure~\ref{fig:sensorDiagram}. Both of the aforementioned sensors
utilized EeonTex LG-SLPA-16K resistive fabric to detect contact with a
robotic arm. Our sensor is designed to be versatile, with the ability
to swap out the resistive material without recreating the entire
sensor from scratch. This is beneficial for experimenting with a wide
variety of materials in order to find the optimal substrate for a
given task. The tactile sensor array designed in~\cite{Day2018} required fewer wires than other designs by using
overlapping rows and columns of conductive fabric on either side of
the resistive substrate. This design, however, was plagued by cross
talk as there was no way to distinguish between signals if multiple
taxels were activated at once. For this reason, we opt for separate
electrodes per taxel in order to achieve more reliable results at the
cost of more wires. The sensor
in~\cite{Bhattarcharjee2013} was designed for large area tactile
sensing covering the entire robotic arm with
taxels ranging in size from 4 to 2 cm$^{2}$. To match our use of sensing on the robot palm we build 1 cm$^{2}$ taxels.
    
For many years, foam has been used in capacitive sensors for microphones~\cite{Sessler2004} and
more recently has made its way into sensing for robotics~\cite{Atalay2018} and sport impact analysis~\cite{Merrell2017}. It is a desirable 
substrate because it is inexpensive and can be tailored to different applications by tuning parameters
such as the thickness and size of the pores to achieve different performance. One intriguing property that was exhibited by piezoresistive
foam constructed out of silicone was significant resistance to drift~\cite{Merrell2013}. Despite these 
factors which make foam a versatile, inexpensive and reliable sensing substrate, to the best of our knowledge
it has never been used in any tactile sensing applications. In this paper, we will show that our novel 
piezoresistive foam sensor is not only comparable to piezoresistive fabrics common in tactile sensing, but that it provides the necessary sensations for both contact detection and object recognition.

%%% Local Variables:
%%% mode: latex
%%% TeX-master: "main"
%%% End:

\section{Tactile Sensor Design}
\label{sec:sensor-design}
%This section will introduce the methodology for constructing the sensor- introducing the materials which will be tested/compared along with design decisions 

% This section will also include the evaluations from experiments such as drift, sensitivity graphs etc
%Subsections:    
            % intro  (not really a section just a segway into the sections)
            %Design - piezoresistive sensor: gnd -> piezoresistive material -> electrode
            %construction methods for fast and scalable skin
                %iron on adhesive, conductive epoxy to connect thread to board
                %use conductive thread rated at 10ohm/cm constant resistance coated in nail polish as wires
        % Comparison of Materials
                %thicknesses, no load resistance, linearity 
                %sensitivity deltaR/R0 graph
                %drift
                %limit of detection
                %response / rise time 
In this section we detail the design and construction of our
sensor. We then provide a comparative analysis between four
piezoresistive substrates which can be used in the sensor.
Importantly, we construct our sensor in such a way to directly compare
different substrates within a common form factor.

% Sensor Design

\subsection{Sensor Design and Construction}
% Overall sensor operation and design
Figure~\ref{fig:sensorDiagram} shows the design of the tactile sensor.
The bottom and top non-conducting fabrics act to electrically insulate the sensor
from the robot and environment. The three layers making up the active
elements of the sensor correspond to a piezoresistive substrate, $\rtax$,
sandwiched between two conductive layers: a lower ground plane and
upper electrode.
By providing a constant
input voltage to the sensor network and connecting an electrode in series with a reference resistor, $\rdiv$, we form a voltage divider that can detect an applied load by measuring the change in output voltage.

% Use soft materials
We follow previous designs~\cite{wang-sensors2019,Bhattarcharjee2013} and construct soft tactile sensors using commonly
available conductive fabric for the two conductive layers of the
sensor. We design and construct our sensor in such a way that we can
directly compare different piezoresistive substrates in identical
sensor form factors. This enables us to compare the use of
piezoresistive foam to the previously used piezoresistive fabrics. 

% Tactile array and electrode operation
As commonly done for tactile sensing, we decompose our sensor
into an array of tactile elements or ``taxels.'' Each taxel
corresponds to an electrode pad made out of conductive fabric.
In order to ease the construction and design of the sensor, all 
taxels in the array share a single ground plane created from a single piece of
conductive fabric. This means that the size and quantity of the taxels
are defined solely by the size and density of the electrodes. A layer of non-conductive
fabric was used to hold the entire sensor together. Each of the electrodes and the ground plane
were attached to the non-conductive fabric layer using an iron-on adhesive. The full list of materials that we used to construct our sensor can be found on our website.

% Ckt Diagram
\begin{figure}[h]
    \vspace{-1mm}
    \centering
    \includegraphics[width=0.46\textwidth]{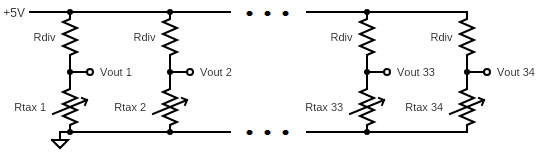}
    \caption{Sensor network diagram for the 34-taxel sensor that was mounted to the Allegro robotic hand.}
    \label{fig:cktDiag}
    \vspace{-5pt}
\end{figure}
The design of our full tactile array can be seen in the sensor network illustrated in
Figure~\ref{fig:cktDiag}. We provide a regulated constant voltage
input of 5V to the network of \(K\) taxels $\rtaxk$, $k=1,\ldots,K$.
To simplify the analysis we elected to use a single value for $\rdiv$ across all substrates, rather than tuning the value for each material. 
 After testing a variety of $\rdiv$ values, we chose a value of 
$ 5.1 \text{k}\Omega$ because it offered the largest full-scale-range for all of the
substrates. The signals from each voltage divider are passed through a
multiplexer to an Arduino Nano that collects and sends each taxel's value to the robot. Since the Arduino Nano has a 10 bit
analog to digital converter (ADC), the digital signal of each taxel is
represented by Eq.~(\ref{eq:1}).
\begin{equation} \label{eq:1}
    ADC = \frac{1024 \rtaxk}{\rdiv + \rtaxk} 
  \end{equation}

\begin{figure}
\vspace{8pt}
    %\begin{turn}{90}
    \includegraphics[trim=0 450 0 100,clip,width=0.46\textwidth]{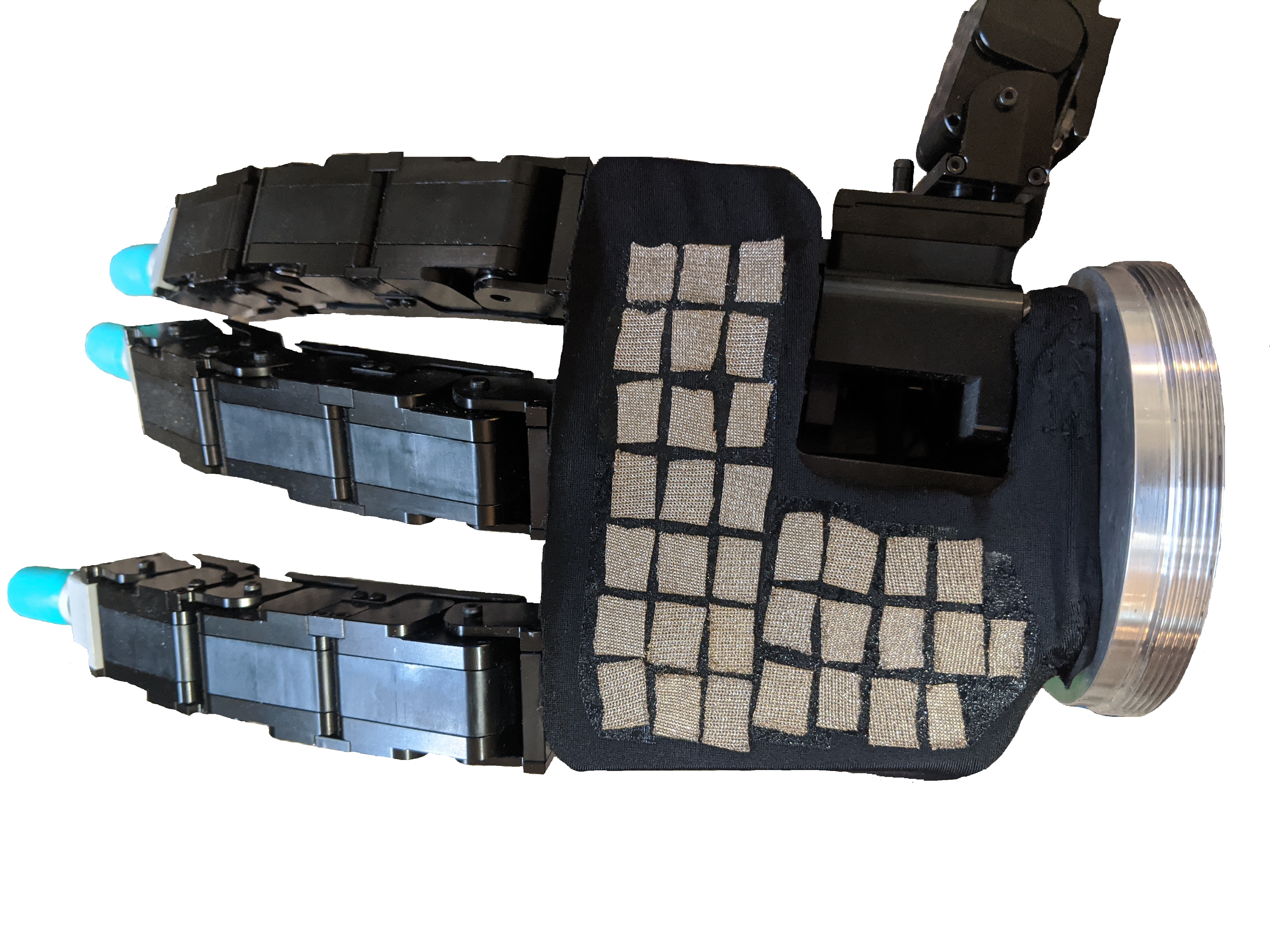}
    %\end{turn}
    \caption{Electrodes mounted to the palm of the Allegro robotic hand.}
    \label{fig:handPic}
    \vspace{-15pt}
  \end{figure}

% Construction
We now describe how we construct our sensor.
Both the electrodes and ground plane are attached to a layer of non-conductive
fabric using an iron on adhesive, a fast and simple
step. Conductive threads coated in an insulating layer of nail polish 
carry the electrodes signals to the rest of the circuit. 
Our sensor is constructed as a pouch such that different substrates
 can be inserted between the ground and electrode planes with
 ease. The pouch is closed at the top with a thin strip of hook and
 loop fastener. While this construction method was ideal for easily switching
 out the substrate for comparisons, it comes at the cost of not having the 
 substrate rigidly attached to the electrode and ground plane. It has been shown 
 that having a stable, rigid connection between the electrode and the substrate
 can improve the signal-to-noise ratio of a sensor \cite{wang2019}. For the benchtop experiments
 outlined in this paper, we secure the electrode to the substrate by lightly sewing them
 together. Even with the pouch structure of our sensor, we were able to obtain clear
 signals sufficient for both contact detection and object recognition. 
 Figure~\ref{fig:handPic} shows the layout of our 34 taxels, each with a 1 cm$^2$ footprint, spaced to cover the palm of the Allgero hand.

 Since the size of each taxel is determined solely by the area of the
 electrodes, taxel areas can be increased or decreased to
 achieve different spatial resolutions at different positions of
 the array. This variable spatial resolution provides the benefits of higher resolution where needed, such as
 along the palm and fingers of the hand, while
 simplifying the wiring and construction in areas where lower spatial
 acuity suffices, such as the back of the hand or the robot arm. 
% Electrode Image
  
\subsection{Sensor Characterization}
We compare four different piezoresistive substrates in the sensor: two
types of fabric and two types of foam. The fabrics used were the EeonTex
NW170SLPA-2k and Wearic
piezoresistive fabric. 
The EeonTex and Wearic fabrics were both designed for use in tactile
sensors. In contrast, both of the foams were designed as anti-static 
packaging for electronic components. One foam we refer to as \emph{high density} 
due to its smaller pores and firmer structure while the other is referred to as
\emph{low density} due to its larger pores and softer structure. Both foams were $1/4$ inch thick
anti static foam. We used a National Instruments 9237 Strain/Bridge Input Module 
with a 10 kHz sampling frequency to record the data referenced in this section. 
% https://www.digikey.com/product-detail/en/desco/12250/16-1231-ND/405476&?gclid=Cj0KCQjwt5zsBRD8ARIsA
%https://www.amazon.com/Multicomp-Density-Non-Corrosive-305x305x6mm-2-Pieces/dp/B071G74PGW

Despite their intended use, the foams demonstrated similar time
responses as the fabrics when a weight was placed and then removed from a single
taxel. The time response plots illustrated in Figure~\ref{fig:riseFallTimeVar} were obtained with a 20g hexagonal weight with a 5.85 cm$^2$
footprint placed in the center of a single 1 inch$^2$ taxel. 
Figure~\ref{fig:riseFallTimeVar} shows the averaged responses of 15
trials for each material with the region within one standard deviation shaded. We measured the rise and fall times as the time it took the signal to change between 10\% and 90\% of its final steady state value.

\begin{figure}[h]
  \vspace{-20pt}
    \centering
        \includegraphics[trim=170 300 185 300,clip,width=0.48\textwidth]{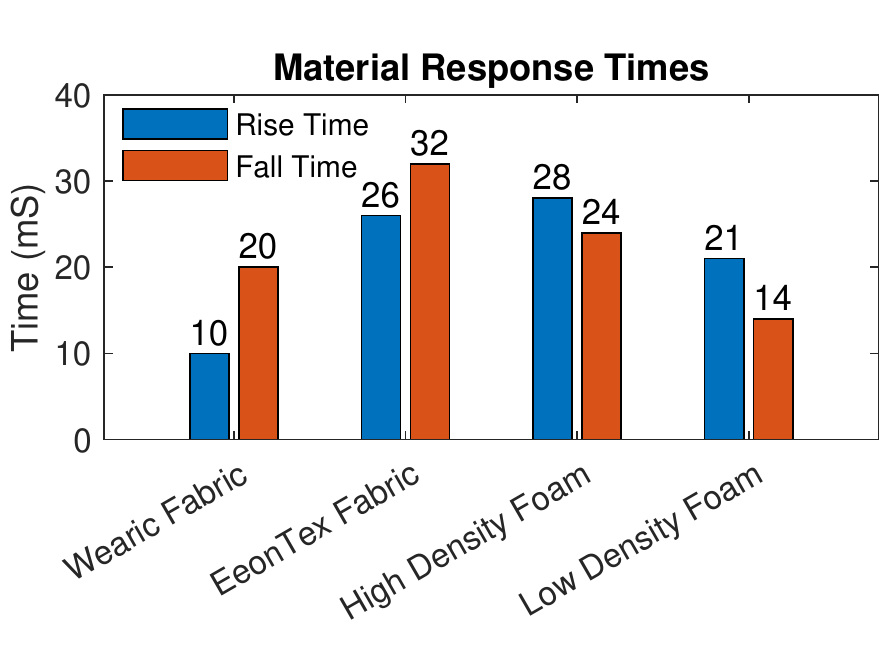}
        \vspace{-8mm}
        \caption{Response time comparison of fabric and foam substrates}
        \label{fig:responseBar}
\end{figure}

\begin{figure}[h]
        \vspace{5pt}
        \includegraphics[trim=110 300 110 300,clip,width=0.48\textwidth]{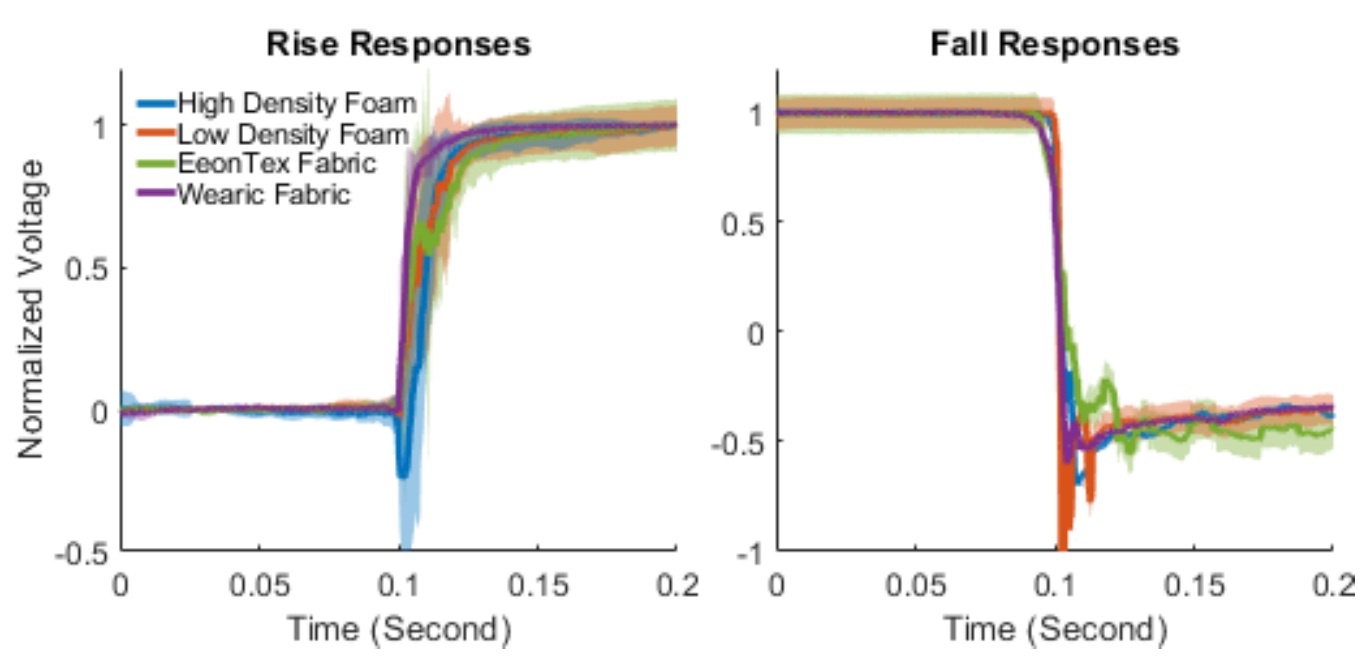}
        \caption{Time response of each material averaged over 15 trials where a 20g weight was placed and then removed from the testing sensor. One standard deviation from the mean has been depicted with the shaded regions on the graph.}
        \label{fig:riseFallTimeVar}
        \vspace{-20pt}
\end{figure}

The foam substrates exhibited similar responses to
the piezoresistive fabrics as well as comparable rise and fall times
as shown in Figure \ref{fig:responseBar}. The high density foam exhibited a 
secondary rise after appearing to have settled, indicating that there was more 
unexpected variations between trials of the high density foam than with other 
materials. This could be due to artifacts from previous trials being exhibited, 
implying that it may be less accurate in detecting forces made in quick succession
compared to the other substrates.  

% \vspace{3mm}
% \begin{table}[t]
% \centering
% \begin{tabular}{p{2.25cm}|p{2cm}|p{2cm}}%|p{1cm}|p{1cm}}
%      Material Type & Rise Time (mS) & Fall Time (mS)\\% & No Load Variance &  Loaded Variance\\ %loaded - 3kPa
%      \hline
%      Wearic Fabric      &  10  & 20 \\% & 1.1e3 & 0.75\\% & 7\\
%      EeonTex Fabric     &  26  & 32  \\%& 287 & 2.3e3\\
%      \hline
%      Low Density Foam   &  21  & 14 \\%& 583 & 145 \\
%      High Density Foam  &  28  & 24 \\%& 442 & 90\\
% \end{tabular}
% \caption{Response time comparison of fabric and foam substrates.}
% \label{table:response-time}
% \vspace{-12pt}
% \end{table}

% Time Response Plots
% \begin{figure}[h]
%   \vspace{-20pt}
%     \centering
%         \includegraphics[trim=140 250 160 250,clip,width=0.48\textwidth]{RiseTime.pdf}
%         \vspace{-8mm}
%         \caption{Time response of each material averaged over 15 trials where a 20g weight was placed onto the testing sensor.}
%         \label{fig:riseTime}
% \end{figure}
% \begin{figure}[h]
%         \includegraphics[trim=140 270 160 275,clip,width=0.48\textwidth]{FallTime.pdf}
%         \caption{Time response of each material averaged over 15 trials where a 20g weight was removed from the testing sensor.}
%         \label{fig:fallTime}
%         \vspace{-20pt}
% \end{figure}
% \begin{figure}[h]
%         \vspace{-10pt}
%         \includegraphics[trim=140 320 145 320,clip,width=0.48\textwidth]{riseFall.pdf}
%         \caption{Time response of each material averaged over 15 trials where a 20g weight was removed from the testing sensor.}
%         \label{fig:riseFallTime}
%         \vspace{-10pt}
% \end{figure}
\begin{figure}
  \vspace{5pt}
        \includegraphics[trim=165 315 160 320,clip,width=0.48\textwidth]{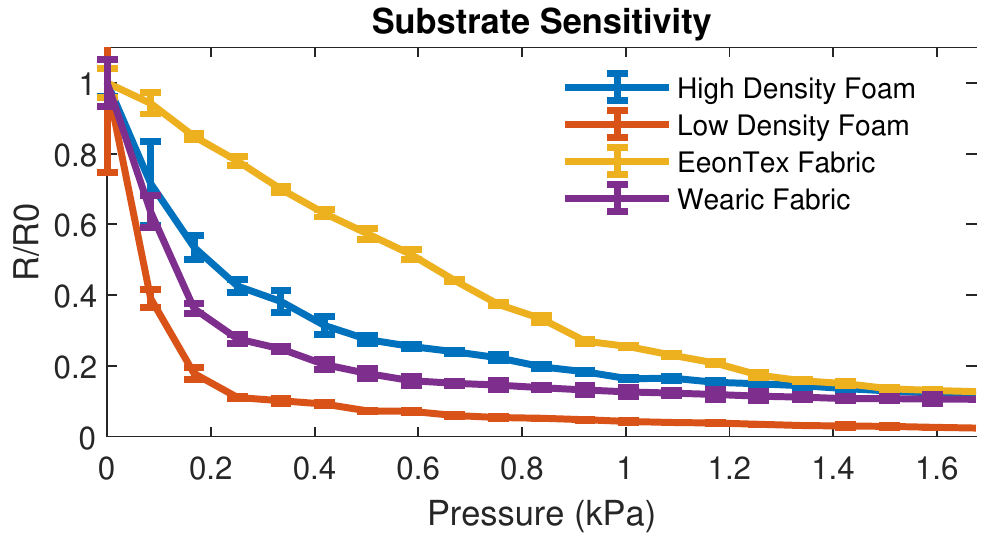}
        %\vspace{-10pt}
        \caption{Comparison of normalized sensitivity of each
          piezoresistive material. For each material, weights from 0 to 100g were placed on a single 8 cm$^2$ taxel in increments of 5g. The weights were placed in the center of the taxel had a footprint of 5.85 cm$^2$. After each weight was placed, voltage was recorded for 5 seconds and then averaged to create a single number per weight measurement.}
        \label{fig:sensitivity}
        \vspace{-20pt}
\end{figure}

In order to determine the sensitivity to changes of pressure for each of the substrates, we recorded data
from a single 1 inch$^2$ taxel for 5 seconds after loading the sensor with a weight. 
Weight was added to the sensor in increments of 5 from 0 to 100g. 
Each time after a new weight was added to the sensor, several seconds were given 
before data began to record to ensure that only steady state data for each weight 
was recorded. The pressure applied to the sensor was then calculated for each weight. Figure \ref{fig:sensitivity} shows the results of this experiment using a total of 1,625 samples from the sensor for each weight increment on 
each of the tested substrates. While the EeonTex fabric had the most linear response, each of the other materials followed
predictable curves. 

As can be seen, the low density foam has a much larger signal change over pressures less than 300Pa, making it well suited for applications where quickly detecting small amounts of force is necessary, such as in contact detection. The high density foam, which is able to detect a larger range of forces before saturating is better equipped for tasks such as object classification. This difference in sensitivity range between the high and low density foams grants researchers much more freedom to fine tune the foam substrate to best fit the needs of their system. Another benefit of using a foam substrate instead of fabric for tactile sensing in robots is that the $1/4$ inch thick foam offers a small amount of mechanical compliance which the fabric does not. This allows the robot a small window of reaction time between initial contact with the foam and contact with the rigid body of the robot. This increases safety by both protecting the robot from collisions with an unmovable object as well as in the case of a collision with a human where the foam would soften the impact significantly.

With a thicker form factor, comes the potential for hysteresis to make it more difficult to detect when contact has occurred. To test this, we recorded the hysteresis for each time the 20g weight was placed and removed from each of the materials. An average of 20 full cycles of placing and removal of the weight were used per material to determine the average hysteresis for each. While we expected to see a higher hysteresis in the foams, they performed very similarly to the fabrics, 17.77\% and 24.14\% hysteresis for the low and high density foams versus  15.15\% and 23.98\% for the Wearic and EeonTex, respectively.

% \begin{table}[t]
% \centering
% \begin{tabular}{p{2.25cm}|r}
%      Material Type & \% Hysteresis\\ 
%      \hline
%      Wearic Fabric      &  15.15\\
%      EeonTex Fabric     &  23.98\\
%      \hline
%      Low Density Foam   &  17.77\\
%      High Density Foam  &  24.14\\
% \end{tabular}
% \caption{Percent hysteresis measured for each tested substrate.}
% \label{table:hysteresis}
% \vspace{-18pt}
% \end{table}

In summary, we find that our easily procurable foams provide results
comparable to previously developed soft, resistive sensors which are
much harder to obtain. 
They also allow researchers more freedom to choose the sensitivity of their substrate to best suit their application. Furthermore while the
foams provided slightly noisier responses and showed higher nonlinear
response across load, we believe the signals are sufficiently predictable for use
in robotics application when coupled with standard signal processing tools.
% Sensitivity Plot
%%% Local Variables:
%%% mode: latex
%%% TeX-master: "main"
%%% End:

\section{Robot Manipulation Experiments}
\label{sec:robot-experiments}
We conduct here demonstration experiments to show the utility of our sensor for use in interactive, robotic manipulation and perception tasks.
We mount the sensor on the four-fingered, 16 DoF Allegro
hand attached to a 7 DoF KUKA LWR4 robot arm.

In our first set of experiments we examine the ability for the sensor
to detect contact fast enough to allow the robot arm to stop before damaging itself or its environment.
Such an ability is paramount for minimizing forces while reaching in clutter~\cite{Killpack2013} or performing 3D object
reconstruction from touch~\cite{yi-iros2016-active-touch}.
In our second set of experiments we examine the utility of our sensor for object and shape recognition.

\subsection{Contact Detection and Motion Arrest Experiments}
For our first experiment we compare the high density and low density foams' ability to robustly detect contact while moving.
We use a simple random forest classifier
trained on sensor data collected from 15 different objects. %  57334 total datapoints in dataset, 25916 in contact points (31418 no contact) 
The data was collected using two different methods with 10 objects selected for each: % grammar? 

\emph{Method 1: } The robot arm approached each object moving in a straight horizontal line. 
Once the sensor made contact with an object, it would push the object until the trajectory was completed. 
For each selected object, this test was repeated five times. 

\emph{Method 2: } The robot arm approached each selected object from a total of five different trajectories. 
Once contact with the object was made, the robot was commanded to halt movement by the press of a button by the experimenters. 
In order to limit the effect of human error, after the data had been recorded, the instant in which contact was made or lost,
in the cases where the object fell over, was refined using the timestamps on images from an RGB-D camera.

The classifier was trained on a total of 57,334 sensor readings. In 45\% of these readings, the sensor was in contact with an object.
For each trial, a baseline reading was selected when the sensor was known to not be in contact.
We zeroed the sensor prior to each experiment by subtracting this baseline signal.
When using the pre-recorded data from the trials described above, the random forest achieved an f1 score of 0.99 % actual 0.9899586
when trained to detect contact and 0.98 % actual 0.9867
when trained to identify which object has been contacted. %These numbers were obtained with 33% test set using the f1 score function in sklearn

% TODO: the following needs to be updated using the new data from the robot experiments 

We conduct a simple motion feedback control experiment to compare the
low and high density foams contact detection performance.
The robot moves in a straight line in the task space from a starting pose towards a goal pose, where an
unknown object sits between the two.
If the robot detects contact using the classifier described above, it commands the controller to stop. We
measure if the sensor detects contact and if so, what happens to the object.

\begin{figure}[h]
  \vspace{-2mm}
  \centering
  \includegraphics[width=0.45\linewidth]{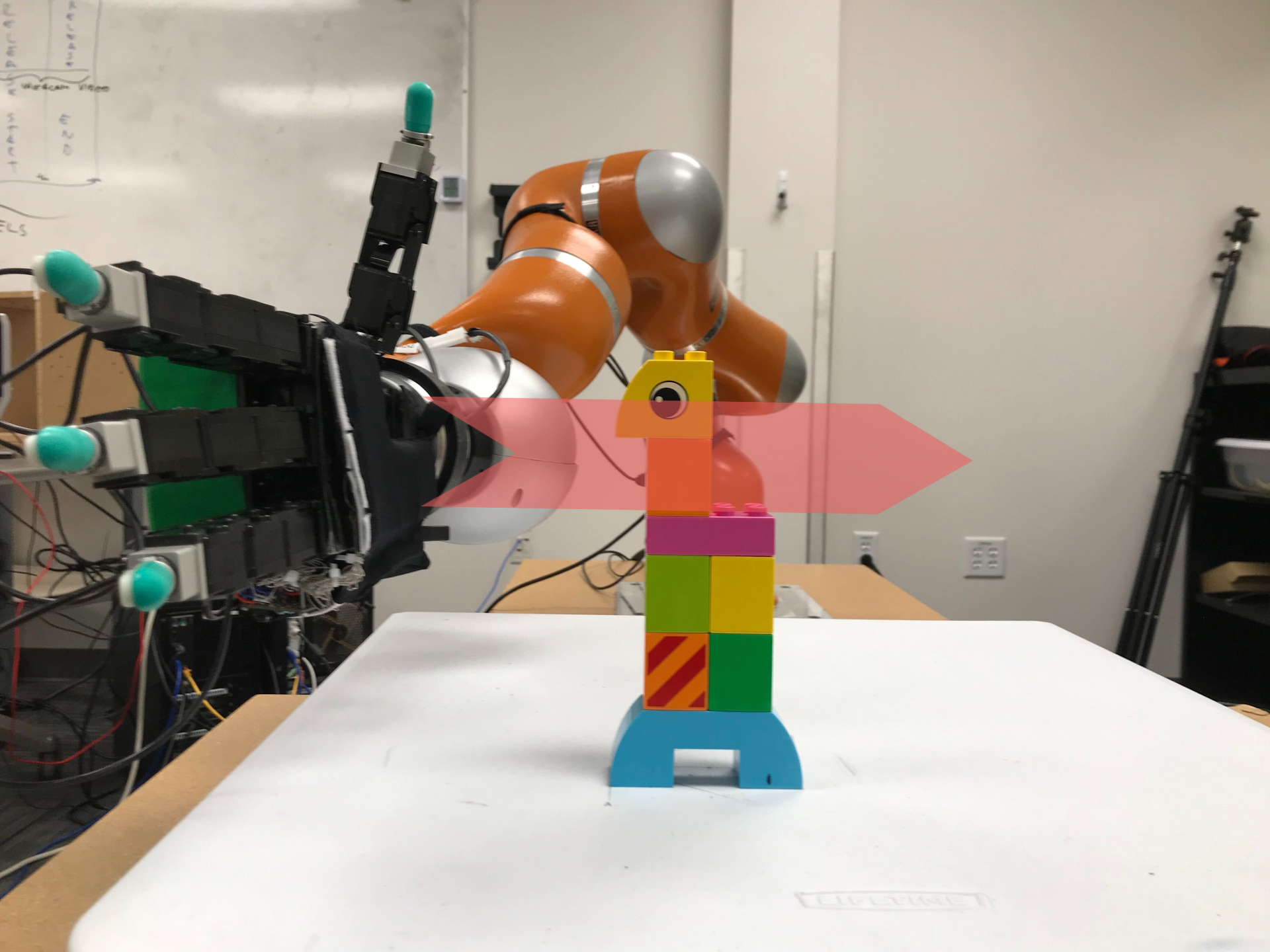}
  \includegraphics[width=0.45\linewidth]{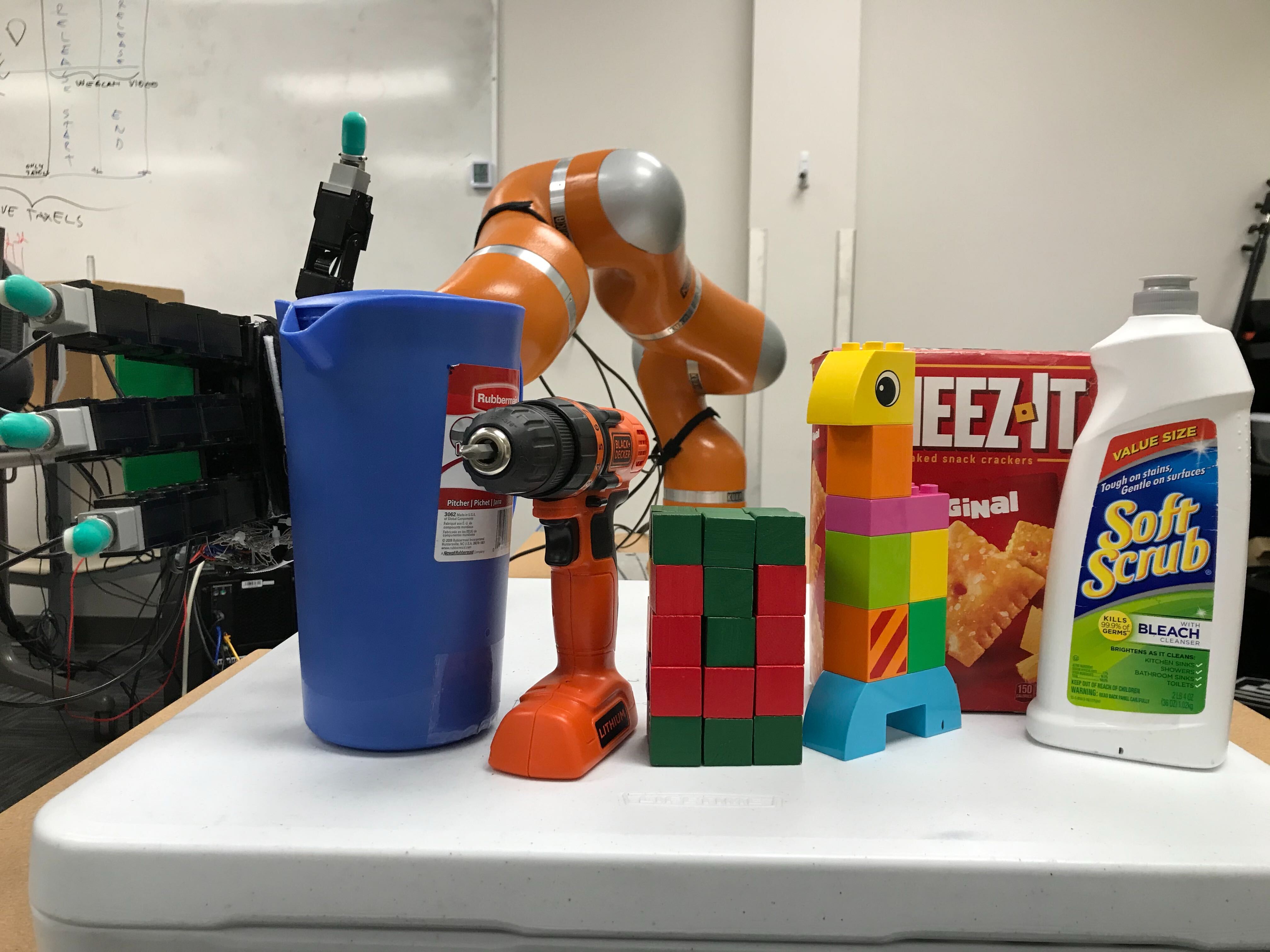}
\caption{Contact detection experimental setup: (left) the arrow indicates
  the robot's motion; (right) objects used: pitcher, drill, \\u-tower, llama, cheezeit, and bleach}
      \label{fig:contact-detection-setup}
\vspace*{-3mm}
\end{figure}

\begin{figure}[h]
    \vspace{-7pt}
    \includegraphics[width=\linewidth,trim=30 0 0 0]{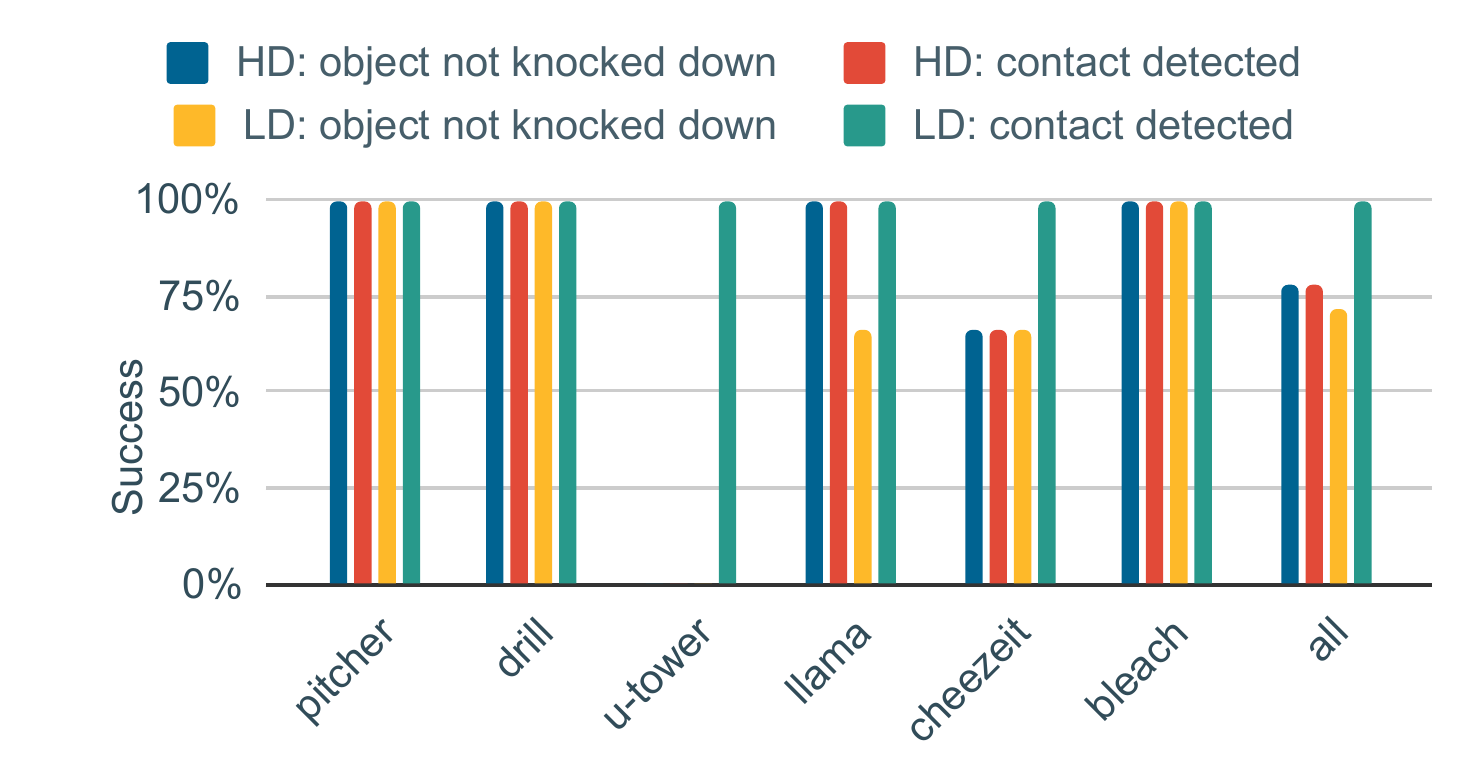}
    \caption{Results comparing high (HD) and low density (LD) foam for
      contact detection.}
    \label{fig:contact-detection-results}
    \vspace*{-6mm}
\end{figure}

We have the robot perform a total of 18 probing actions for each foam across 6 different objects of differing geometry. 
Figure~\ref{fig:contact-detection-setup} shows the experimental setup and the 6 test objects.
We place each object in the same location, 12 cm from the starting pose.
We perform one trial for each of three different orientations per
object to test the sensor's detection ability against different local
geometries. The results are presented in Figure~\ref{fig:contact-detection-results}.
Videos of all experiments can be found at our website.

We find that the \textbf{low density foam detected contact in every
  trial}, while the high density foam detected contact in 14 of the 18
trials. The high density foam never detected contact with the \emph{u-tower},
and failed in one trial with the \emph{cheezit} box. The
\emph{u-tower} is constructed from a set of non-rigidly attached
wooden cubes. Thus it loads only a minor force on the sensor prior to falling apart, 
which our simple classifier has difficulty detecting. Similarly, a contact with \emph{cheezeit} isn't registered at some orientations where only a slight force is sufficient to knock it over.

The results show that both substrates can detect contact fast enough to not
knock down the heavier objects. We believe learning a more sophisticated
contact classifier from labeled data
(e.g.~\cite{veiga-toh2018-slip-prediction})
 as well as rigidly mounting the electrodes to the substrate as was demonstrated in 
\cite{wang-sensors2019} could significantly improve the quality of contact detection for both substrates.

\subsection{Object Recognition Experiments}
\begin{figure}[h]
    \vspace{-7pt}
    \includegraphics[width=\linewidth]{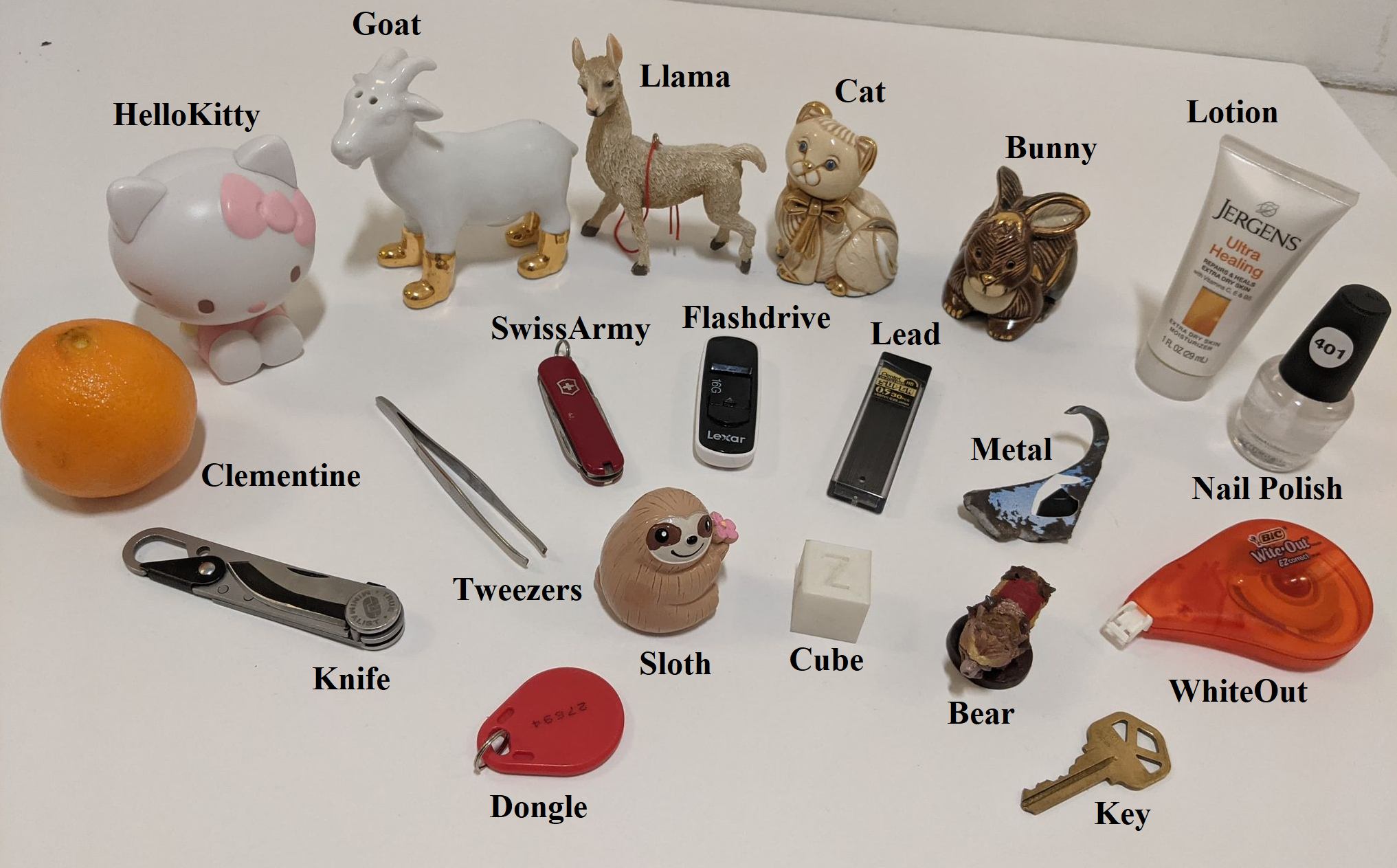}
    \caption{Objects used for object detection using a 20-taxel sensor.}
    \label{fig:object-detection-items}
    %\vspace*{-8mm}
\end{figure}

Based on the findings of the contact detection experiments, we 
selected to further evaluate the low density foam in the context of 
object recognition and contact localization. In order to further measure the effectiveness of the sensors ability to distinguish between a variety of forces, we performed object recognition on a set of 20 unique objects with a 5x4 rectangular taxel array with the low density foam as the sensing substrate. The objects that were used for this experiment can be seen in Fig.~\ref{fig:object-detection-items}. 

We performed a total of three data collection trials between
which the order of the objects was randomized. This was
done in order to avoid any artifacts left on the sensor from
heavier items such as the cat and bunny which each weighed more than 100 grams.
Each object was sampled in multiple orientations throughout each trial. The
objects were picked up and replaced into the center of the
sensor array for each data point. This granted a large variety
of samples from every angle of each object to be stored into the
dataset. In the end, over the 20 different objects we collected
a total of 1172 samples.

\begin{figure}[h]
    \vspace{8pt}
    \includegraphics[width=\linewidth,clip,trim=1.5cm 0cm 2.7cm 3mm]{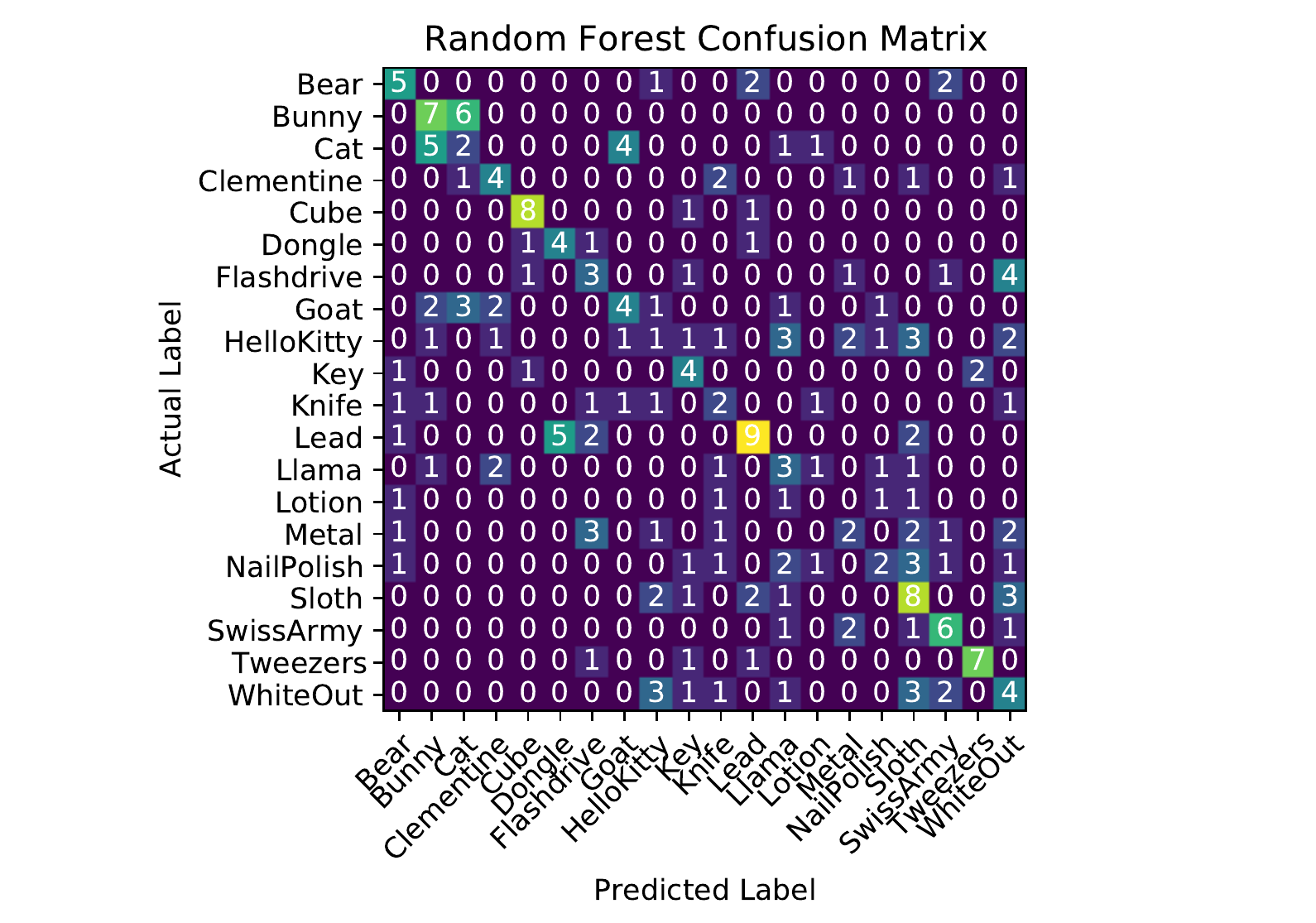} % trim order l b r t
    \caption{Confusion matrix achieved using a random forest classifier attempting to classify 20 different objects with a 20-taxel sensor.}
    \label{fig:object-detection-confusion}
    \vspace*{-7mm}
  \end{figure}

We trained a random forest classifier from scikit-learn\cite{scikit-learn} on this data with a 80/20 train test split. Using this method, we achieved 36.2\% accuracy and a confusion matrix depicted in Fig.~\ref{fig:object-detection-confusion}. While this number may seem low, recall that randomly guessing only provides a 5\% chance of guessing the correct object, so learning is taking place. Further, as can be seen in the confusion matrix, objects with similar weights and sizes were difficult to distinguish such as the bunny and cat or the dongle and lead objects.

% \begin{figure}[h]
%     \vspace{-8pt}
%     \includegraphics[width=\linewidth,clip,trim=0cm 1cm 0cm 3mm]{forest_39_warm.png} % trim order l b r t
%     \caption{Confusion matrix achieved using a random forest classifier attempting to classify 20 different objects with a 20 taxel sensor.}
%     \label{fig:object-detection-confusion}
%     \vspace*{-6mm}
%   \end{figure}
One approach to overcome the limitation in performance shown here is to leverage the change in contact information over time during a grasping process.
To do so we examined the tactile signals present during object lifting when using a state-of-the-art grasp planner~\cite{lu-iros2020-active-grasp} to generate grasps
for unknown objects with only partially observed geometry from an RGB-D camera. 

\begin{figure}[h]
   % \vspace{-2mm}
    \includegraphics[width=.24\textwidth]{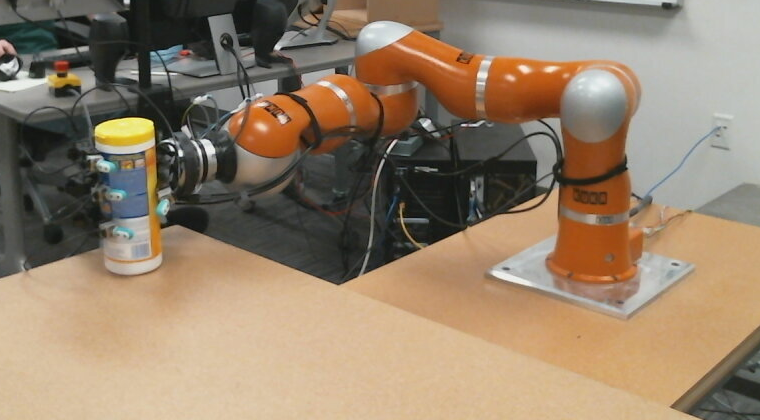}\hfill
    \includegraphics[width=.24\textwidth]{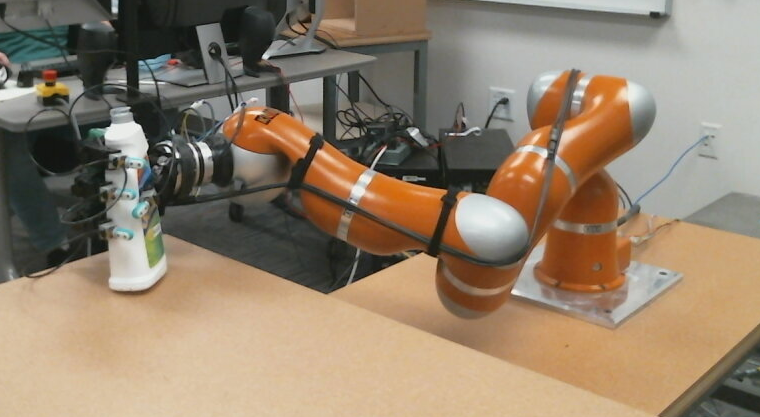}
    \\[\smallskipamount]
    \includegraphics[width=.24\textwidth]{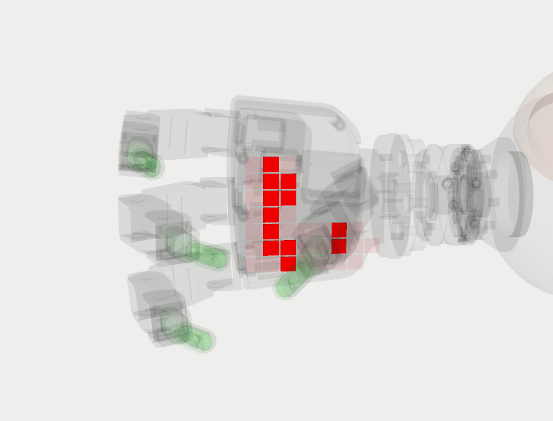}\hfill
    \includegraphics[width=.24\textwidth]{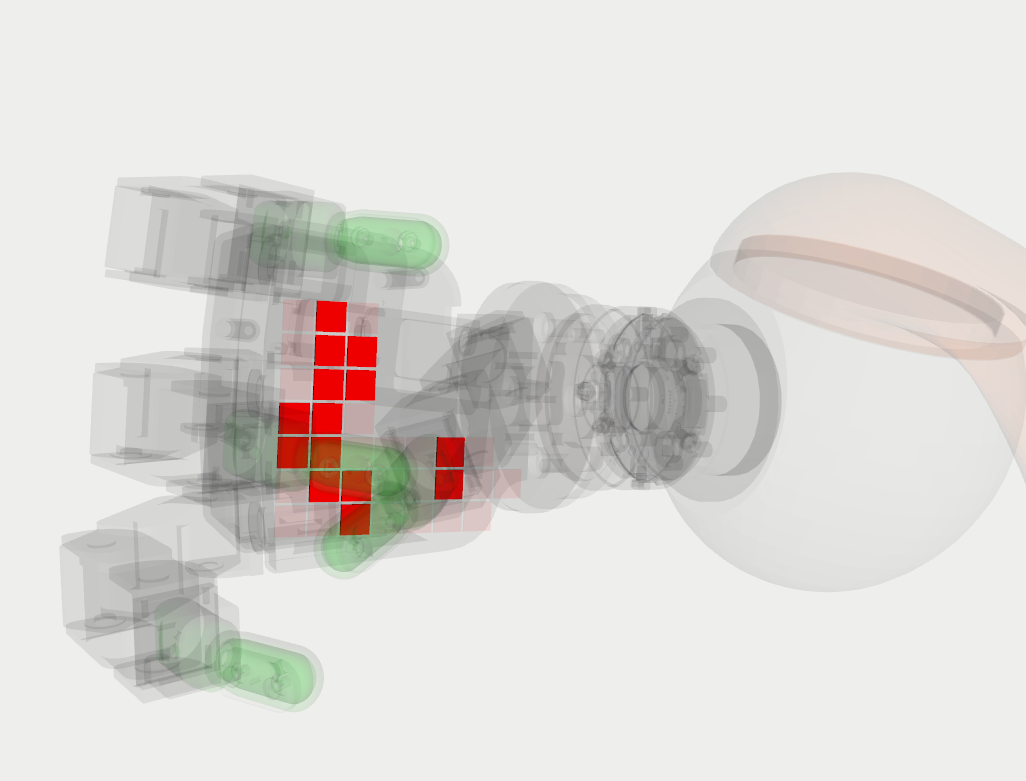}
\caption{Different taxels activate depending on the object shape and
  pose in the hand.}
\label{fig:grasp-qualitative}
\vspace*{-3mm}
\end{figure}

We can see in Figure~1 that a different set of taxels activates during
grasping and lifting of the same object. We attribute this to the
object shifting once it loses the support of the table.
This change in contact over time could be leveraged for increased object recognition performance.
This also indicates the sensor's ability to detect in-hand pose change,
which could likely enable slip detection if significant changes continue to occur. 
We highlight additional results in Fig.~\ref{fig:grasp-qualitative}. The
distinct contact patterns for different objects provide further support for
the sensor's ability to conduct shape reconstruction or object detection. 

%%% Local Variables:
%%% mode: latex
%%% TeX-master: "main"
%%% End:

%%% Local Variables:
%%% mode: latex
%%% TeX-master: "main"
%%% End:

\section{Conclusion}
\label{sec:conclusion}
In this paper we evaluated the performance of two types of
piezoresistive foam for use in tactile sensing. We compare the
performance of these foams in controlled experiments to two 
piezoresistive fabrics specifically designed for tactile
sensing applications. Our results show that the foams provided
similar responses under load to the fabrics, albeit with somewhat
higher levels of noise. Given that these foams were not designed for
use in sensing, but instead for electronic protection, we find
these results extremely promising.

Our sensor was constructed with the piezoresistive substrate in a 34-taxel sensor array,
mounted to the palm of an Allegro robotic hand. We tested the performance of
the sensors to detect and localize contact, as well as analyze the
performance of autonomous grasp executions. We found the sensor to
provide useful and meaningful response, enabling the robot to stop
fast enough to not knock over objects it made contact with and provide
distinct signatures corresponding to different grasping configurations.

These results show that foam has a significant potential for use in
tactile skins. These skins could both be applied on hands for
manipulation as in this work or to cover robot arms to improve
reaching in clutter or operating around humans as previously shown
with piezoresistive fabric
sensors~\cite{Bhattarcharjee2013,Killpack2013,Day2018}.
We note that
the mechanical compliance offered by the foam can likely be leveraged to offer
additional safety for the robot and its surrounding including human
users and compatriots.

There are numerous advantages for utilizing foam instead of fabric as 
a sensing substrate, one of which is being easy to source. Piezoresistive
fabric is only sold by a small handful of manufacturers. The primary 
manufacturer of piezoresistive fabric, Eeonyx, %makers of the EeonTex fabric we used 
has discontinued production of it which has made it even more difficult
to source. By contrast, foam is an extremely common material which can be 
readily purchased from many different sources. Recent work has indicated that 
any open-cell polyurethane foam can be converted into a reliable piezoresistive
sensor by applying a coating of conductive ink~\cite{wang-sensors2019}. This greatly 
increases the variety of foams which can be purchased and converted into sensors.

Our experiments, coupled with this ability to easily transform any
foam into a suitable substrate, opens the door to much exciting future work in developing a piezoresistive foam that is optimized for tactile sensing applications. This includes reducing hysteresis and drift observed by the foam sensors. Due to its porous nature, foam has been shown to be very successful in detecting vibrations~\cite{Sessler2004}. This means that there is a potential for foam substrates to detect slip and other tactile events. 
Finally, we note that this increased spatial resolution of tactile
sensing on hands can enable improved grasp analysis, including automatic detection and learning of grasp type~\cite{lu-ral2019-grasp-types}.

%%% Local Variables:
%%% mode: latex
%%% TeX-master: "main"
%%% End:

\section*{Acknowledgments}
{\footnotesize
Rebecca Miles was supported in part by NSF Award \#1841845. Martin
Matak was supported by NSF Award \#1846341 and by DARPA under grant
N66001-19-2-4035. Mohanraj Devendran Shanthi was supported by DARPA
under grant N66001-19-2-4035.}
% Add this temporarily for easier examination of space.
%\clearpage\newpage
%% Use plainnat to work nicely with natbib.

\bibliographystyle{ieeetr}
{\footnotesize
\bibliography{references}
}
\end{document}